# Separable and Transitive Graphoids

Dan Geiger*
Northrop Research and
Technology Center
One Research Park
Palos Verdes, CA 90274

David Heckerman†
Medical Computer Science Group
Knowledge Systems Laboratory
Departments of Computer Science and
Medicine, Stanford University, CA 94305

## Abstract

We examine the notion of "unrelatedness" in a probabilistic framework. Three formulations are presented. In the first formulation, two variables $a$ and $b$ are *totally independent* with respect to a set of variables $U$ if they are independent given any value of the variables in $U$. In the second formulation, two variables are *totally uncoupled* if $U$ can be partitioned into two marginally independent sets containing $a$ and $b$ respectively. In the third formulation, two variables are *totally disconnected* if the corresponding nodes are disconnected in any belief network representation. We explore the relationship between these three definitions of "unrelatedness" and explain their relevance to the process of acquiring probabilistic knowledge from human experts.

## 1 Introduction

The notion of dependence between variables plays a key role in the theory of belief networks and in the way they are used for inference. The intuition that guides the construction of these networks draws from the analogy between "connectedness" in graphical representations and "dependence" in the domain represented, that is, two nodes connected along some path correspond to variables which are dependent in some way. Below concepts of relatedness are examined and a theoretical foundation for this intuition is provided.

We examine three formulations of the sentence "$a$ and $b$ are totally unrelated". In the first formulation, two variables $a$ and $b$ are said to be *totally independent* iff they are independent given any value of the other variables in the domain, namely, if $U$ is a set of variables and $P$ is a probability distribution, then $a$ and $b$ are totally independent wrt $P$ iff

$$P(a,b|Z=\boldsymbol{Z}) = P(a|Z=\boldsymbol{Z}) \cdot P(b|Z=\boldsymbol{Z})$$

where $Z$ is any subset of $U$ not containing $a$ and $b$, and $\boldsymbol{Z}$ is any value of $Z$. In the second formulation, two variables are said to be *totally uncoupled* wrt $P$ if $U$ can be partitioned into two marginally independent sets $U_1$ and $U_2$ containing $a$ and $b$ respectively, namely,

$$P(U_1=\boldsymbol{U_1}, U_2=\boldsymbol{U_2}) = P(U_1=\boldsymbol{U_1}) \cdot P(U_2=\boldsymbol{U_2})$$

where $U_1 \cup U_2 = U$, $U_1 \cap U_2 = \emptyset$, and $\boldsymbol{U_1}, \boldsymbol{U_2}$ are any values of these variables. In the third formulation, two variables $a$ and $b$ are *totally disconnected* if the corresponding nodes are disconnected in every belief network representation of $P$.

Are these three formulations equivalent and if not, under what conditions do they coincide? Our main contribution is to identify a class of distributions called *transitive* for which all three formulations are equivalent. Strictly positive binary distributions and regular Gaussian distributions (defined below) are examples of transitive distributions. We also show that "connectedness" in graphical representations and "dependence" in the domain represented (i.e., the converse of total independence) are equivalent for every transitive distribution and for none other.

These results have several theoretical and practical ramifications. Our analysis uses a qualitative abstraction of probabilistic independence known as graphoids (Pearl and Paz, 1989) and it demonstrates the need for this abstraction in manipulating independence assumptions (which are an integral part of every probabilistic reasoning engine). Our proof also demonstrates that belief networks provide a powerful mathematical tool for understanding of probability theory itself. Finally, we demonstrate the rele-vance of these results to simplifying the process of acquiring probabilistic knowledge from experts.

*Supported in part by NSF GRANT IRI-8821444, while the author was at UCLA.

†Supported by the National Library in Medicine GRANT RO1LM04529, and by NSF GRANT IRI-8703710.

## 2 Separability and Transitivity

Throughout our discussion we consider a finite set of variables $U = \{u_1, ..., u_n\}$ each of which is associated with a finite *set of values* $d(u_i)$ and a probability distribution $P$ with the Cartesian product $\prod_{u_i \in U} d(u_i)$ as its sample space. We use lowercase letters possibly subscripted (e.g $a$, $b$, $x$ or $u_i$) to denote variables, and use uppercase letters (e.g. $X$, $Y$, or $Z$) to denote sets of variables. A bold lowercase or uppercase letter refers to a value of a variable or set of variables, respectively. A value $\boldsymbol{X}$ of a set of variables $X$ is a member in the Cartesian product $\prod_{x \in X} d(x)$ where $d(x)$ is the set of values of $x$. The notation $X = \boldsymbol{X}$ stands for $x_1 = \boldsymbol{x}_1, ..., x_n = \boldsymbol{x}_n$ where $X = \{x_1, ..., x_n\}$ and $\boldsymbol{x}_i$ is a value of $x_i$.

**Definition** Let $U = \{u_1, ..., u_n\}$ be a finite set of variables with $d(u_i)$ and $P$ as above. If $X$, $Y$, and $Z$ are three disjoint subsets of $U$, then $X$ is *probabilistically independent* of $Y$ given $Z$, denoted $I_\mathcal{P}(X, Y; Z)$, iff for every three sets of values $\boldsymbol{X}$, $\boldsymbol{Y}$, and $\boldsymbol{Z}$ of $X, Y$, and $Z$, respectively, the following equation holds:

$$P(X = \boldsymbol{X}, Y = \boldsymbol{Y} | Z = \boldsymbol{Z}) = P(X = \boldsymbol{X} | Z = \boldsymbol{Z}) \cdot P(Y = \boldsymbol{Y} | Z = \boldsymbol{Z}) \quad (1)$$

Every probability distribution induces a dependency model:

**Definition** (Pearl, 1988) A *dependency model* $M$ over a finite set of elements $U$ is any set of triplets $(X, Y; Z)$ where $X$, $Y$ and $Z$ are disjoint subsets of $U$. The intended interpretation of $M$ is that $(X, Y; Z) \in M$ iff $X$ is *independent* of $Y$ given $Z$, also denoted by $I(X, Y; Z)$.

A probability distribution induces a dependency model when we identify $I$ with $I_\mathcal{P}$.

When speaking about dependency models, we use both set notations and logic notations. If $(X, Y; Z) \in M$, we say that the *independence statement* $I(X, Y; Z)$ holds for $M$. Similarly, we either say that $M$ contains a triplet $(X, Y; Z)$ or that $M$ satisfies a statement $I(X, Y; Z)$. An independence statement $I(X, Y; Z)$ is called an *independency* and its negation is called a *dependency*.

Graphoids are special types of dependency models:

**Definition** (Pearl and Paz, 1989) A *graphoid* is any dependency model $M$ which is closed under the following *axioms*[1]:

Trivial Independence

$$I(X, \emptyset; Z) \quad (2)$$

Symmetry

$$I(X, Y; Z) \Rightarrow I(Y, X; Z) \quad (3)$$

Decomposition

$$I(X, Y \cup W; Z) \Rightarrow I(X, Y; Z) \quad (4)$$

Weak union

$$I(X, Y \cup W; Z) \Rightarrow I(X, Y; Z \cup W) \quad (5)$$

Contraction

$$I(X, Y; Z) \,\&\, I(X, W; Z \cup Y) \Rightarrow \\ I(X, Y \cup W; Z) \quad (6)$$

It can readily be shown that probabilistic independence ($I_\mathcal{P}$) satisfies these axioms, and therefore every probability distribution defines a graphoid. Several additional types of graphoids are discussed in (Pearl, 1988; Pearl and Paz, 1989). A simple example of graphoids is given below. Consider a graphoid $M_1$ over $U = \{a, b, c, d\}$ which consists of the independence statement $I(\{a, b\}, \{c, d\}; \emptyset)$ and the ones derivable from it by the graphoid axioms. Notice that using weak union, decomposition, and symmetry axioms, $I(\{a, b\}, \{c, d\}; \emptyset)$ implies that the following statements are in $M_1$ as well:

$$\{I(a, c; \emptyset),\ I(a, c; b),\ I(a, c; d),\ I(a, c; \{b, d\})\}$$

and, therefore, $a$ and $c$ are totally independent. Similarly, $b$ and $c$ are totally independent.

Next we define total independence and total uncoupledness in graphoid terminology.

**Definition** Let $M$ be a graphoid over a finite set of elements $U$. Two elements $a$ and $b$ of $U$ are said to be *totally independent* (wrt $M$) iff $(a, b; Z) \in M$ for every subset $Z$ of $U \setminus \{a, b\}$. When $a$ and $b$ are not totally independent, then we say that $a$ and $b$ *interact* and denote it by $interact(a, b)$.

**Definition** Let $M$ be a graphoid over a finite set of elements $U$. Two elements $a$ and $b$ of $U$ are said to be *totally uncoupled* (wrt $M$) iff there exist two subsets $U_1$ and $U_2$ of $U$ such that $a \in U_1$, $b \in U_2$, $U_1 \cap U_2 = \emptyset$, $U_1 \cup U_2 = U$, and $(U_1, U_2; \emptyset) \in M$. When $a$ and $b$ are not totally uncoupled, then we say that $a$ and $b$ are *coupled*.

---
[1]This definition differs slightly from that given in (Pearl and Paz, 1989) where axioms (3) through (6) define semi-graphoids. Axiom (2) is added for clarity.

Notice that due to symmetry, decomposition, and weak union axioms total uncoupledness implies total independence. The converse does not always hold. For example, if $U$ consists of three variables $a$, $b$ and $c$, then it is possible that $a$ and $b$ are totally independent, namely that $a$ and $b$ are both marginally independent [i.e. $I(a,b;\emptyset)$], and independent conditioned on $c$, and yet no variable is independent of the other two. This happens, for example, if $a$ and $b$ are the outcome of two independent fair coins and $c$ is a variable whose domain is $\{head, tail\} \times \{head, tail\}$ and whose value is $(i,j)$ if and only if the outcome of $a$ is $i$ and the outcome of $b$ is $j$.

This example leads to the following definition:

**Definition** A graphoid $M$ over a finite set of elements $U$ is *separable* iff every two totally independent elements $a$ and $b$ are totally uncoupled.

The property of separability, as it turns out, can be cast in another appealing format; it is equivalent to the requirement that interaction (the converse of total independence) is transitive, namely, that interact satisfies axiom 7 below.

**Definition** A graphoid $M$ over a finite set of elements $U$ is *transitive* iff

$$interact(a,b) \ \& \ interact(b,c) \Rightarrow$$
$$interact(a,c) \qquad (7)$$

This axiom is so appealing to our intuition that we are tempted to speculate that all distributions not obeying this property are *epistemologically inadequate* for modeling a human reasoner, and that distributions that do satisfy this property are *natural* in the sense that they adequately represent the conventional properties of the word "interact".

The following theorem establishes the equivalence between separability and transitivity. The proofs of this and subsequent theorems can be found in the appendix.

**Theorem 1** *A graphoid $M$ over a finite set of elements is separable iff it is transitive.*

Probabilistic independence between sets of variables, as is well known, is not determined by the independencies between their individual elements; two sets may be dependent although their individual elements are independent. For example, if $a$ and $b$ represent the outcome of two independent fair coins (using 0's and 1's) and $c$ is their sum modulo 2 then $c$ is dependent on $\{a,b\}$ but is independent of each individual variable. The absence of this compositional property stands in contrast to our intuition because we normally expect that a proposition unrelated to the pieces of some body of knowledge is unrelated to the whole as well.

Lemma 2 below shows that total independence does not suffer from this anomaly.

**Definition** Let $M$ be a graphoid over a finite set of elements $U$. Two disjoint subsets $A$ and $B$ of $U$ are *totally independent* wrt $M$ iff $(A, B; Z) \in M$ for every $Z$ that is a subset of $U \setminus A \cup B$.

**Lemma 2** *Let $M$ be a graphoid over a finite set of elements $U$ and let $A$, $B$, and $C$ be three subsets of $U$. If $A$ and $B$ are totally independent and $A$ and $C$ are totally independent then, $A$ and $B \cup C$ are totally independent as well.*

**Proof:** Denote the sentence "$X$ is totally independent of $Y$" with $J(X,Y)$. By definition, $J(A,B)$ implies $(A,B;Z) \in M$ and $J(A,C)$ implies $(A,C;Z \cup B) \in M$ where $Z$ is an arbitrary subset of $U \setminus A \cup B \cup C$. Together, these statements imply by contraction that $(A, B \cup C; Z) \in M$. Hence, $J(A, B \cup C)$ holds. $\square$

The definition of total uncoupledness can similarly be extended to sets and it also satisfies the compositional property stated by Lemma 2.

## 3 Connectedness in Belief Networks

We have introduced two ways of defining totally unrelated elements of a graphoid; total independence, and total uncoupledness. Here we suggest a third approach—total disconnectedness: Elements $a$ and $b$ are *totally disconnected* in $M$ if they are disconnected in every belief network representation of $M$. We shall see below that two elements are totally disconnected in $M$ if and only if they are totally uncoupled in $M$.

**Definition** Let $M$ be a graphoid over a finite set of elements $U$. A directed acyclic graph $D$ is a *belief network representing $M$* iff there exists a one to one mapping between elements in $U$ and nodes of $D$, and $D$ is constructed from $M$ by the following steps: assign a total ordering $d : u_1, u_2, ,..., u_n$ to the elements of $U$. For each element $u_i$ in $U$, identify a minimal set of predecessors $\pi(u_i)$ such that

$(u_i, \{u_1, ..., u_{i-1}\} \setminus \pi(u_i); \pi(u_i)) \in M$. Assign a direct link from every node corresponding to an element in $\pi(u_i)$ to the node corresponding to $u_i$.

For example, a belief network representing $M_1$, our example from the previous section, constructed in the order $a$, $b$, $c$, and $d$ consists of four nodes $a$, $b$, $c$ and $d$ and edges from $a$ to $b$ and from $c$ to $d$. Another belief network of $M_1$, constructed in the order $d$, $c$, $b$, and $a$ yields a graph with reversed edges. In general, different orderings yield networks with different sets of edges.

**Definition** A *trail* in a belief network is a sequence of links that form a path in the underlying undirected graph. Two nodes are *connected* in a belief network $D$ iff there exists a trail connecting them in $D$. Otherwise they are disconnected. A *connected component* of a belief network $D$ is a subgraph $D'$ in which every two nodes are connected, and $D'$ is *maximal* iff there exists no supergraph of it with this property.

**Definition** Two elements of a graphoid $M$ are said to be *totally disconnected* iff in every belief network representation of $M$ the nodes corresponding to these elements are disconnected. Otherwise these elements are *connected* in $M$.

For example in $M_1$, $a$ and $c$ are totally disconnected. However, to verify this fact would have been quite difficult without the next theorem which shows that it suffices to examine a single belief network representation of $M_1$ in order to determine disconnectedness, rather than to check all such representations.

**Theorem 3** *Two elements of a graphoid $M$ are disconnected in some belief network representation of $M$ iff they are disconnected in every belief network representation of $M$ (i.e., disconnected in $M$).*

Consequently, we obtain:

**Theorem 4** *Let $M$ be a graphoid. Two elements $a$ and $b$ of $M$ are totally disconnected iff they are totally uncoupled.*

We have thus far obtained the relationships between three formulations of unrelatedness: total disconnectedness and total uncoupledness are equivalent, both are stronger than total independence, and all three definitions are equivalent for transitive (separable) graphoids.

## 4 Separable Distributions and Instantiated Graphoids

The notion of separability developed so far would have remained unrealized unless examples of separable graphoids were provided. This section provides such examples. Our plan is to introduce a new axiom, propositional transitivity, show that it implies separability and that it holds for regular Gaussian distribution and strictly positive binary distributions. Consequently, these type of distributions are separable.

**Definition** A *regular Gaussian* distribution is a multivariate normal distribution with finite nonzero variances and with finite means. A *strictly-positive binary* distribution is a probability distribution where every variable has a domain of two values, say 0 and 1, and every combination of the variables' values has a probability greater than zero.

The definition of dependency models and graphoids of section 2 precludes the representation of statements of the form "$a$ and $b$ are independent given $c = c_1$, yet dependent given $c = c_2$" because graphoids do not distinguish between values of a variable. Thus in order to represent axioms that refer to specific values of a variable (as propositional transitivity does), requires a slight modification of these definitions.

**Definition** Let $U$ be a finite set of variables each associated with a finite set of values. An *instantiated dependency model* $M$ over $U$ is a set of triplets $(\boldsymbol{X}, \boldsymbol{Y}; \boldsymbol{Z})$ where $X$, $Y$ and $Z$ are disjoint subsets of $U$, and $\boldsymbol{X}$, $\boldsymbol{Y}$ and $\boldsymbol{Z}$ are their values respectively.

Clearly, every instantiated dependency model $M_R$, defines a dependency model $M$ in the sense of section 2; a triplet $(X, Y; Z)$ is in $M$ iff $(\boldsymbol{X}, \boldsymbol{Y}; \boldsymbol{Z})$ is in $M_R$ for every value $\boldsymbol{X}, \boldsymbol{Y}, \boldsymbol{Z}$ of $X$, $Y$ and $Z$, respectively. The model $M_R$ is said to *induce* $M$. In particular, every probability distribution defines an instantiated dependency model.

**Definition** An *instantiated graphoid* is any instantiated dependency model that induces a graphoid.

**Theorem 5** (*Geiger and Heckerman, 1989*)
*Regular Gaussian distributions and strictly positive binary distributions satisfy the following axiom, named* propositional transitivity:[2]

$$I(A_1A_2A_3A_4, B_1B_2B_3B_4; \emptyset) \,\&$$
$$I(A_1A_2B_3B_4, B_1B_2A_3A_4; e = e') \,\&$$

---

[2]In complicated expressions, $A_1A_2$ is used as a shorthand notation for $A_1 \cup A_2$ and $eA_1$ denotes $\{e\} \cup A_1$.

$$I(A_1A_3B_2B_4, B_1B_3A_2A_4; e = \boldsymbol{e''}) \Rightarrow$$
$$I(A_1, eA_2A_3A_4B_1B_2B_3B_4; \emptyset) \vee$$
$$I(B_1, eA_1A_2A_3A_4B_2B_3B_4; \emptyset) \qquad (8)$$

*Where all sets mentioned are pairwise disjoint and do not contain $e$, and $\boldsymbol{e'}$ and $\boldsymbol{e''}$ are distinct values of $e$.*

**Theorem 6** *Every instantiated graphoid satisfying propositional transitivity is separable.*

Regular Gaussian distributions satisfy axioms other than propositional transitivity which can be used to show separability. A particularly interesting one is *unification*:

$$I(X = \boldsymbol{X}, Y = \boldsymbol{Y}; Z = \boldsymbol{Z}) \Rightarrow I(X, Y; Z)$$

which states that if $X$ and $Y$ are independent given one arbitrary value of $X$, $Y$, and $Z$, then these sets are independent given every value of $Z$. Thus, although, regular Gaussian distributions have infinite domains, the number of independencies is finite and can be completely represented assuming finite domains.

We have chosen, however, to focus on propositional transitivity because this choice allows us to unify the separability proof for two quite different classes of distributions, thus, demonstrating the power of this axiomatic approach.

We conjecture that propositional transitivity holds also for binary distributions that are not strictly positive.

## 5 Probabilistic Knowledge Acquisition

The construction of belief networks as faithful representations of a given domain relies on the ease and confidence by which an expert can describe the relationships between variables in this domain. Explicating these relationships is often straightforward but may encounter difficulties when variables have many values. For example, in medical diagnosis, a variable corresponding to "cancer" may have dozens of possible values, each corresponding to a different type of cancer. An expert wishing to describe the relationship between the different symptoms, tests, and treatments of cancer may find it rather confusing unless he first partitions the many types of cancer into several groups sharing common characteristics; in fact, the grouping of related pieces of information into more or less independent chunks is an important step in organizing any large body of knowledge. Below, we shall see how the theory developed in previous sections facilitates this task, through the construction of *similarity networks* (Heckerman, 1990a; Heckerman, 1990b).

Let $h$ be a distinguished variable designated for the disease "hypothesis", and let the values of $h$, $\boldsymbol{h}_1$, ..., $\boldsymbol{h}_n$, stand for an exhaustive list of possible diseases. First, a connected undirected graph is constructed where each of the $n$ nodes represents a different value of $h$ and each link represents a pair of "similar" diseases, namely diseases that are sometimes hard to discriminate. Then, for each link $\boldsymbol{h}_i$—$\boldsymbol{h}_j$ in the graph, a *local belief network* is composed, assuming that either $h = \boldsymbol{h}_i$ or $h = \boldsymbol{h}_j$ must hold; it consists of a distinguished root node $h$, additional nodes that are *connected* to $h$ representing symptoms, and links representing the dependencies among these symptoms and their relationship to the hypothesis node $h$. Finally, the global network is formed from the local networks; it consists of the union of all links and their adjacent nodes in the local networks.

In (Heckerman, 1990b), it is shown that under the assumption of strict-positiveness, namely that every combination of symptoms and diseases is feasible, the union of the connected components of node $h$ in each local network generates a belief network that faithfully represents the domain. That is, the assertions of conditional independence encoded in the graph union of the local networks are logically implied by the assertions of conditional independence in each of the local networks. Although when using this methodology we must construct many local networks instead of one, there are two important advantages to such a composition. First, local networks for pairs of similar diseases tend to be small. Second, by composing local networks for pairs of similar diseases, the expert can direct his attention on those diagnostic subproblems with which he is familiar and thereby increase the quality of the knowledge he provides.

A difficulty with this approach is to identify the set of nodes that are connected to node $h$ in each local network. In principle, we could consult the expert by asking him directly queries of the form: "is node $s$ ($s$ connotes symptom) connected to node $h$, given that either $h = \boldsymbol{h}_i$ or $h = \boldsymbol{h}_j$ must hold?". This query, however, may be inadequate because it refers to a graphical representation of the domain, a language with which the expert might not be familiar. On the other hand, the query "does this symp-tom in any circumstances help you to discriminate between the two diseases $\boldsymbol{h}_i$ and $\boldsymbol{h}_j$ ?" is much more appealing since it addresses directly the knowledge of the expert.

The first query asks about total disconnectedness of $s$ and $h$, while the second query, which is concentrated on determining whether $P(s|\boldsymbol{h}_i, \boldsymbol{Z}) \neq P(s|\boldsymbol{h}_j, \boldsymbol{Z})$ for some values of some set of variables $Z$, corresponds to asking about total independence of $s$ and $h$. This paper shows that total disconnectedness and total independence coincide for transitive distributions and identifies important classes of distributions that are transitive. Consequently, using the second query in the construction of similarity networks is a theoretically-justified heuristic; and indeed its soundness has been empirically verified (Heckerman, 1990b).

## 6 Summary

We have examined the notion of unrelatedness of variables in a probabilistic framework. We have introduced three formulations for unrelatedness—total independence, total uncoupledness, and total disconnectedness— and explored their interrelationships. ¿From a practical view point, these results legitimize prevailing decomposition techniques of knowledge acquisition; it allows an expert to decompose the construction of a complex belief network into a set of belief networks of manageable size.

Our proof technique uses the qualitative notion of independence as captured by the axioms of graphoids and would have been much more difficult had we used the probabilistic definitions of conditional independence. This axiomatic approach enables us to identify a common property—propositional transitivity—shared by two distinct classes of probability distributions (regular Gaussian and strictly-positive binary), and to use this property without attending to the detailed characteristics of the classes.

In addition, we have shown that useful classes of probability distributions are transitive, the proof of which is facilitated greatly by the network formulation. Thus, we see that network representations, apart of their dominant role in representing experts' opinions, are also a powerful mathematical tool for uncovering formal properties of independence relationships.

## Appendix

Some preliminary definitions are needed.

**Definition** (Pearl, 1988) A node $b$ is called a *head-to-head* node wrt a trail $t$ iff there exist two consecutive edges $a \rightarrow b$ and $b \leftarrow c$ on $t$. A trail $t$ is *active by $Z$* if (1) every head-to-head node wrt $t$ either is or has a descendent in $Z$ and (2) every other node along $t$ is outside $Z$. Otherwise, the trail is said to be *blocked by $Z$*.

**Definition** (Pearl, 1988) If $X$, $Y$, and $Z$ are three disjoint subsets of nodes in a dag $D$, then $Z$ is said to *d-separate* $X$ from $Y$, denoted $I_\mathcal{D}(X, Y; Z)$, iff there exists no active trail by $Z$ between a node in $X$ and a node in $Y$.

The next theorem states that $d$-separation is a sound criteria; in (Geiger et al., 1990) $d$-separation is shown to be complete as well. These results are fundamental to the theory of belief networks.

**Theorem 7** (Verma and Pearl, 1988) *Let $D$ be a belief network representing a graphoid $M$ over a finite set of variables $U$. If $X$ and $Y$ are d-separated by $Z$ in $D$ then $(X, Y; Z) \in M$.*

**Lemma 8** *Let $D$ be a belief network representing a graphoid $M$ over a finite set of variables $U$. If $A$ and $B$ are two subsets of $U$ and they correspond to two disconnected sets of nodes in $D$, then $(A, B; \emptyset) \in M$.*

**Proof:** Follows directly from the theorem above (7); there is no *active* trail between a node in $A$ and a node in $B$ which makes $A$ and $B$ $d$-separated given the empty set, hence, $(A, B; \emptyset)$ is in $M$. □

**Proof of Theorem 3:** It suffices to show that any two belief networks representing $M$ share the same maximal connected components. Let $D_A$ and $D_B$ be two belief networks representing $M$. Let $C_A$ and $C_B$ be maximal connected components of $D_A$ and $D_B$ respectively. Let $A$ and $B$ be the nodes of $C_A$ and $C_B$ respectively. We show that either $A = B$ or $A \cap B = \emptyset$. This will complete the proof because for an arbitrary maximal connected component $C_A$ in $D_A$ there must exists a maximal connected component in $D_B$ that shares at least one node with $C_A$ and thus, by the above claim, it must have exactly the same nodes as $C_A$. Thus each maximal connected component of $D_A$ shares the same nodes with exactly one maximal connected component of $D_B$. Hence $D_A$ and $D_B$ share the same maximal connected components.

Since $D_A$ is a belief network representing $M$ and $C_A$ is a connected component of $D_A$, by Lemma 8,

$(A, U \setminus A; \emptyset) \in M$, where $U$ stands for $M$'s elements. By symmetry (3) and decomposition (4), $(A \cap B, B \setminus A; \emptyset) \in M$. A subgraph is a connected component only if its nodes cannot be partitioned into two sets $U_1$ and $U_2$ such that $(U_1, U_2; \emptyset) \in M$. Hence, for $C_B$ to be connected either $A \cap B$ or $B \setminus A$ must be empty. Similarly for $C_A$ to be connected, $A \cap B$ or $A \setminus B$ must be empty. Thus, either $A = B$ or $A \cap B = \emptyset$. □

**Proof of Theorem 4:** Assume $a$ and $b$ are disconnected in one belief network. Then $(U_1, U_2; \emptyset) \in M$ where $U_1$ are the elements corresponding to nodes connected to $a$ and $U_2$ are the rest of $M$'s elements (Lemma 8). Hence $a$ and $b$ are uncoupled.

Assume $a$ and $b$ are uncoupled, and that $U_1$ and $U_2$ are the two independent sets which partition $M$'s elements and which contain $a$ and $b$ respectively. Constructing a belief network for $M$ in an order in which all elements in $U_1$ are placed before any element in $U_2$ yields a network where $U_1$ and $U_2$ are disconnected. □

**Proof of Theorem 1:** First we notice that if $M$ is separable, namely, when total independence implies total uncoupledness, then $M$ is transitive because in this case total uncoupledness coincides with total independence and coupledness is transitive.

It remains to show the converse; transitivity implies separability. Let $U$ stand for $M$'s elements. Let $a$ and $b$ be two arbitrary elements of $U$ that do not interact. We will show by induction on $|U|$ that if *interact* satisfies transitivity (7) then there exists a belief network representation $D$ of $M$ where $a$ and $b$ are disconnected. Consequently, $a$ and $b$ are uncoupled (Theorem 4) and, therefore, $M$ is separable.

We construct $D$ in the ordering $u_1 \equiv a, u_2 \equiv b, u_3, ..., u_n \equiv e$ of $M$'s elements. Assume $n = 2$. Variables $a$ and $b$ do not interact, therefore $(a, b; \emptyset) \in M$. Thus, $a$ and $b$ are not connected. Otherwise, $n > 2$. Let $D_e$ be a belief network formed from $M$ by the ordering $u_1, ..., u_{n-1}$ of $M$'s elements. let $A$ be the set of nodes connected to $a$ and let $B$ be the rest of the nodes in $D_e$. The network $D$ is formed from $D_e$ by adding the last node $e$ as a sink and letting its parents be a minimal set that makes $e$ independent of all the rest of the elements in $M$ (see the definition of belief networks). By the induction hypothesis, before $e$ was added, $A$ and $B$ are disconnected. After node $e$ is added, a trail through $e$ might exists that connects a node in $A$ and a node in $B$. We will show that there is none; if the parent set of $e$ is indeed minimal, then either $e$ has no parents in $A$ or it has no parents in $B$, rendering $a$ and $b$ disconnected.

Since $a$ and $b$ do not interact and since $M$ satisfies transitivity (7), it follows that either $a$, or $b$, do not interact with $e$. Without loss of generality assume that $a$ and $e$ do not interact. Let $a'$ be an arbitrary node in $A$. By transitivity it follows that either $a$ or $e$ do not interact with $a'$, for otherwise, $a$ and $e$ would interact, contrary to our selection of $a$. If $a$ and $a'$ do not interact, then by the induction hypothesis, $A$ can be partitioned into two independent subsets, thus $A$ would not be connected in the belief network $D_e$, contradicting our selection of $A$. Thus, every element $a' \in A$ does not interact with $e$. It follows that the entire set $A$ does not interact with $e$ (Lemma 2). Thus, in particular, $(A, e; \hat{B}) \in M$ where $\hat{B}$ are the parents of $e$ in $B$. Consequently, $e$ has no parents in $A$ because otherwise its set of parents were not minimal because $\hat{B}$ would make $e$ independent of all other elements in $U$. Hence, $a$ and $b$ are on two different connected components of $D$. □

The following definition which abstracts the notion of conditional distribution, is needed for the next proof.

**Definition** Let $M(U)$ be an instantiated dependency model over a finite set of variables $U = \{u_1, ..., u_n\}$. The *conditional of $M(U)$ on $u_n = \boldsymbol{u}_n$*, denoted $M(\{u_1, ..., u_{n-1}\}|u_n = \boldsymbol{u}_n)$, is an instantiated dependency model that contains a triplet $(\boldsymbol{X}, \boldsymbol{Y}; \boldsymbol{Z})$ iff $(\boldsymbol{X}, \boldsymbol{Y}; \boldsymbol{Z} \cup \{\boldsymbol{u}_n\}) \in M(U)$

**Proof of Theorem 6:** Let $M$ be an instantiated graphoid and let $U$ be its variables. Let $a$ and $b$ be two arbitrary variables in $U$ that are totally independent. We will show by induction on $|U|$ that if $M$ satisfies propositional transitivity then there exists a belief network representation $D$ of $M$ where $a$ and $b$ are disconnected. Consequently, $a$ and $b$ are uncoupled (Theorem 4) and, therefore, $M$ is separable.

We construct $D$ in the ordering $u_1 \equiv a, u_2 \equiv b, u_3, ..., u_n \equiv e$ of $M$'s variables. Assume $n = 2$. Variables $a$ and $b$ are totally independent, therefore, $(a, b; \emptyset) \in M$. Thus, $a$ and $b$ are disconnected. Otherwise, $n > 2$. Let $D_e$ be a belief network formed from $M$ by the ordering $u_1, ..., u_{n-1}$ of $M$'s variables. let $A$ be the set of nodes connected to $a$ and let $B$ be the rest of the nodes in $D_e$. Since $a$ and $b$ are totally independent, by the induction hypothesis, $(A, B; \emptyset) \in M$ ($\equiv I_1$). The network $D$ is formed from $D_e$ by adding the last node $e$ as a sink and letting its parents be a minimal set that makes $e$ independent of the rest of $M$'s variables (see the definition of belief Networks). Let $D_{\boldsymbol{e'}}$ and

$D_{e''}$ be belief networks of the conditional graphoids $M(A \cup B | e = \text{e'}) (\equiv M_{e'})$ and $M(A \cup B | e = \text{e''}) (\equiv M_{e''})$ respectively, formed in the ordering $u_1, ..., u_{n-1}$. Since both $M_{e'}$ and $M_{e''}$ are subsets of $M$, $a$ and $b$ are totally independent in both these dependency models. By the induction hypothesis $M_{e'}$ and $M_{e''}$ are separable. Hence there exists a partitioning $A_{e'}, \hat{A}_{e'}, B_{e'}$ and $\hat{B}_{e'}$ of $A \cup B$ where $A = A_{e'}\hat{A}_{e'}$, $B = B_{e'}\hat{B}_{e'}$, $a \in A_{e'}$ and $b \in B_{e'}$, such that $(A_{e'}\hat{B}_{e'}, B_{e'}\hat{A}_{e'}; \emptyset) \in M_{e'}$ ($\equiv I_2$). Similarly, there exists a possibly-different partitioning $A_{e''}, \hat{A}_{e''}, B_{e''}$ and $\hat{B}_{e''}$ of $A \cup B$ where $A = A_{e''}\hat{A}_{e''}$, $B = B_{e''}\hat{B}_{e''}$, $a \in A_{e''}$ and $b \in B_{e''}$, such that $(A_{e''}\hat{B}_{e''}, B_{e''}\hat{A}_{e''}; \emptyset) \in M_{e''}$ ($\equiv I_3$). In other words, each of the two instances of $e$ induces a partitioning of $A$ and $B$ into two independent subsets. There are at most eight disjoint subsets formed by the two partitioning. These are: $A_1 \equiv A_{e'} \cap A_{e''}$, $A_2 \equiv \hat{A}_{e'} \cap A_{e''}$, $A_3 \equiv A_{e'} \cap \hat{A}_{e''}$, $A_4 \equiv \hat{A}_{e'} \cap \hat{A}_{e''}$, $B_1 \equiv B_{e'} \cap B_{e''}$, $B_2 \equiv \hat{B}_{e'} \cap B_{e''}$, $B_3 \equiv B_{e'} \cap \hat{B}_{e''}$ and $B_4 \equiv \hat{B}_{e'} \cap \hat{B}_{e''}$. These definitions yield the following relationships: $A = A_1 A_2 A_3 A_4$, $A_{e'} = A_1 A_3$, $\hat{A}_{e'} = A_2 A_4$, $A_{e''} = A_1 A_2$, $\hat{A}_{e''} = A_3 A_4$, $B = B_1 B_2 B_3 B_4$, $B_{e'} = B_1 B_3$, $\hat{B}_{e'} = B_2 B_4$, $B_{e''} = B_1 B_2$ and $\hat{B}_{e''} = B_3 B_4$. Rewriting assertions $I_1, I_2$ and $I_3$ using these notations yields $(A_1 A_2 A_3 A_4, B_1 B_2 B_3 B_4; \emptyset) \in M$, $(A_1 A_3 B_2 B_4, B_1 B_3 A_2 A_4; e = \text{e'}) \in M$ and $(A_1 A_2 B_3 B_4, B_1 B_2 A_3 A_4; e = \text{e''}) \in M$ which are the three antecedents of propositional transitivity (8). Since $M$ is closed under this axiom, it follows that either $(A_1, eA_2 A_3 A_4 B_1 B_2 B_3 B_4; \emptyset) \in M$ or $(B_1, eA_1 A_2 A_3 A_4 B_2 B_3 B_4; \emptyset) \in M$. Since $a \in A_1$ and $b \in B_1$, $M$ is separable. $\square$

## Acknowledgments

This research is based on our dissertations (Geiger, 1990; Heckerman, 1990b). We thank Judea Pearl for encouraging us to pursue the subject, Dan Hunter for checking our proofs, and Jeff Barnett, Norman Dalkey, Ross Shachter, Steve Smith, and Tom Verma for helpful comments.